\documentclass[final,5p,times,twocolumn,dvipsnames]{elsarticle}

\usepackage{amsmath,amssymb}  % For mathematical symbols
\usepackage{graphicx}         % For including graphics
\usepackage{hyperref}         % For hyperlinks
\usepackage{xcolor,here}
\usepackage{lineno}           % For line numbers (optional)
\usepackage{multicol}         % For multi-column formatting
\usepackage{tikz}
\usepackage{booktabs} % For \toprule, \midrule, \bottomrule
\usetikzlibrary{shapes,arrows,calc}

\newcommand{\e}{\vskip 0.5mm \ \\}

\usepackage{amsmath,amssymb,amsfonts, xcolor, bm,here}
\usepackage{algorithm,amssymb,amsmath}
\usepackage{algpseudocode}
\usepackage{soul,xcolor}
\usepackage{multirow}
\usepackage{booktabs}
\usepackage{graphicx}
\usepackage{sidecap} 
\usepackage{siunitx}
\usepackage{textcomp}
 \usepackage{tikz}
\usetikzlibrary{arrows,shapes,calc}
\usepackage{pgfplots}
\usepackage{subcaption}
\usetikzlibrary{positioning}
\newtheorem{rem}{\sc Remark}

\newtheorem{definition}{\sc Definition}

\journal{Control Engineering Practice}

\begin{document}

\begin{frontmatter}

% Title
\title{Identifying Explicit Parsimonious Piece-wise Polynomial Relationships in Industrial time-series: Application to manipulator robots}

% Authors and Affiliations
\author[gipsa]{Mazen Alamir\corref{cor1}}
\ead{mazen.alamir@grenoble-inp.fr}
\author[staubli,gipsa]{Sacha Clavel}
\ead{s.clavel@staubli.com}

\cortext[cor1]{Corresponding author}

% Affiliations
\address[gipsa]{Univ. Grenoble Alpes, CNRS, Grenoble INP, GIPSA-lab, 38000 Grenoble, France}
\address[staubli]{Staubli company. Auvergne-Rhône-Alpes, 74210,  Faverges, France.}

% Abstract
\begin{abstract}
This paper addresses the problem of identifying parsimonious explicit piece-wise polynomial relationships that might involve a relatively large number of raw features. The algorithm leverages a recently proposed identification algorithm that yields parsimonious implicit relationships enabling to derive normality characterization in the context of anomaly detection and localization. The algorithm proposed in this paper goes a step further by deriving \textbf{explicit} piece-wise representations that are built using the set of polynomials involved in the implicit  representations. The framework is illustrated on the problem of identifying parsimonious explicit representations of the inverse model of a  6-axis manipulator robot. Moreover, further experiments on a 4-axis robot are also shown which are designed to investigate the generalization capability of parsimonious models compared to state-of-the-art DNNs structures, when models face unseen contexts of use. 
\end{abstract}

% Keywords
\end{frontmatter}

\section{Introduction}\label{sec:introduction}
\noindent The identification of general nonlinear relationships \cite{alma991007429591406161,aguirre2022birdseyeviewnonlinear} from available sensors data is a key issue in many engineering problems such as, building digital twins, control design, state estimation, virtual sensors design and anomaly detection, to cite but a few. It is no surprise that this topic has been investigated since a long time ago and that the problem remains open as far as general solutions that perform  on any use-case is concerned.  
\e 
The burst of Deep Neural Networks (DNNs) in all their variants \cite{cho2014learningphraserepresentationsusing, alma991007429591406161, vaswani2023attentionneed} distillates the \textit{inaccurate} belief according to which, the problem is behind us. The later belief is based on the universal approximation property of DNNs and the now available parallel computational devices. This suggests  that one has \textit{just}  to collect the necessary amount of data and then to wait \textit{enough time} for the gradient descent process to converge to a \textit{sufficiently good} solution for the resulting model to \textit{miraculously} emerge. 
\e Unfortunately, the specificity of industrial time-series tends to temper this \textit{optimism} \cite{alma991007851304106161, AuffarthBen2021MLfT, Audibert_2022, Rewicki_2023}. Indeed, the industrial context brings additional difficulties and set of challenging requirements that make the problem hard to tackle in a totally satisfactory way. Some of these specificities related to industrial context are discussed in the following section.
\subsection{Specificities of the industrial context} 
\noindent {\sc \color{MidnightBlue} Training datasets are almost never complete}.\\
Indeed, industrial equipments operate within a high variety of contexts of use that are generally not all present in the training datasets. This increases the risk of overfitting even if several regularization techniques such as early stopping can be implemented to prevent the curse of overfitting. While this risk affects all models, it is particularly present in DNNs because of the generally high number of parameters involved. 
\e {\sc \color{MidnightBlue} Explainability is mandatory}. \\ Indeed, prior to their consent to use data-driven models \textit{in production}, industrial operators need to have a minimum understanding of the structure of relationships and the underlying coupling that should validate their own physically-based knowledge of the equipments. In particular when anomalies are detected, it is particularly important in the industrial context to be able to provide a root-cause analysis of the alarm. To this respect, DNN models involving tens, if not, hundreds of thousands of parameters obviously do not address this concern unless a specific a posteriori sensitivity analysis is undertaken which remains a tedious task with generally not rigorously assessed outcomes despite recent meaningful advances \cite{KelekoAurelienTeguede2023Hcmo,KelekoTeguedeAurelien2022DLfa,BobekSzymon2025TFdf}.
\e {\sc \color{MidnightBlue} Industrial relationships are inherently parsimonious}. \\ 
Industrial equipments are designed based on a set of physical laws which are structurally parsimonious. This makes algorithms that are designed to capture this parsimony a priori more appropriate as this makes them less sensitive to the specificity of the training data and hence less prone to bad generalization on unseen contexts. 
\e 
This contribution is based on the belief that the parsimony is an answer to all of the above-mentioned specific aspects of industrial concerns. Indeed: 
\e 
1) Parsimoniously identified relationships are more closely related to the true \textit{hidden} physical relationships and are hence more robust to unseen contexts over which they are more likely to remain valid. 
\e 
2) Parsimonious relationships are structurally built upon a priori sensitivity analysis and hence provide a priori more explainable structures through a battery of relationships (residual generators), each involving a strict subset of arguments that are mandatory to making the relationship stand. This enables a clear diagnosis procedure by pointing out \textit{which sensors} are involved in the residual that goes beyond the alarm's threshold. 
\e 
3) Finally, it goes without saying that if the relationships are inherently parsimonious, then parsimonious identification is the natural target to pursue. 
\subsection{This contribution}
\noindent The present paper leverages a recently proposed algorithm \cite{AlamirMazen2025Aafs} that identifies a set of multi-variate polynomials $P_\kappa$, with $\kappa=1,\dots,n_m$ such that the residual defined by:
\begin{equation}
e(x,y) := \min_{\kappa=1}^{n_m}\Bigl\vert y - P_\kappa(x)\Bigr\vert \label{defdeimplicite}
\end{equation}
is sufficiently small with a moderate number ($n_m$)  of polynomials. In other words, at each instant, the label $y$ is close to one of the predicted values $P_\kappa(x)$ provided by the set of $n_m$ polynomials. 
\e In \cite{AlamirMazen2025Aafs}, the precision of the so-derived implicit relationship has been compared to a DNN structure in terms of residuals amplitude, computation time and complexity. The comparison showed orders of magnitude of difference in terms of complexity and computation time for far lower residuals values. Notice however that, in the case of DNN, the residual is defined by:
\begin{equation}
e(x,y) = \vert \hat y(x)-y\vert \qquad \hat y(x)= \texttt{DNN}(x)\label{defdeeex}
\end{equation}
where  $\hat y=\texttt{DNN}(x)$ is an explicit prediction of the label as a function of the vector of features $x$.
\e Consequently, while the comparison stands when the task is to perform anomaly detection as in \cite{AlamirMazen2025Aafs}, the fact that the latter model provides an explicit prediction of the label while the former \textit{only} produces an implicit  residual renders the two approaches not fully comparable. For instance, when the task is to derive virtual sensors replacing the physical sensor $y$ by the virtual sensors $\texttt{DNN}(x)$, or when $\texttt{DNN}(x)$ is used as a feedforward term in a control loop, the \textbf{implicit} piece-wise polynomial approach does not provide an alternative.
\e
The present contribution fills the gap by deriving parsimonious \textbf{explicit} piece-wise polynomial representations of the form:
\begin{equation}
\hat y:= \dfrac{1}{n_V}\sum_{\ell=1}^{n_V} P_{\kappa^\star_\ell(x)}(x)\quad \text{where}\quad \kappa^\star_\ell: \mathbb R^n\rightarrow \{1,\dots,n_m\}\label{newexplicit}
\end{equation}
in which, the set of polynomials $\{P_\kappa\}_{\kappa=1}^{n_m}$ are those identified for the implicit residual generation relationship \eqref{defdeimplicite} while each of the $n_V$ maps $\kappa^\star_\ell$ defines a specific partition of the features space that maps sub-regions of $\mathbb R^n$ into the set of indices $\{1,\dots,n_m\}$ where the associated polynomial applies. 
\e 
More precisely, a number $n_V$ of different \textit{voters} (predictors) are designed as shown later. The voter with index $\ell\in \{1,\dots,n_V\}$ uses the map $\kappa_\ell^\star$ so that, at feature $x$, it predicts the label using $P_{\kappa^\star_\ell(x)}(x)$ thus mapping to the most suitable polynomial according to its own partitioning. The overall explicit prediction is then taken to be the average of the predictions by the different voters. 
\e Based on the above presentation, the contribution of the present paper is threefold: 
\begin{itemize}
\item[$\checkmark$] The derivation of the heuristics leading to the above partitioning maps $\kappa_\ell^\star$ and consequently to a set of predictors to be averaged.
\item[$\checkmark$] The application of the resulting explicit piece-wise polynomial identification to a real-life data associated to a 6-axis manipulator robots shown in Figure \ref{fig-staubliTx290}. This can be viewed as a solid \textit{proof of concept} for the proposed solution. The results are compared to some state-of-the art DNN structures.
\item[$\checkmark$] In addition, additional dedicated experiments are provided with a 4-axis manipulator where truly different contexts of use are considered for training and test in order to examine and compare the generalization capabilities of the different architectures. 
\end{itemize}
The paper is organized as follows: \e 
Section \ref{sec-defnot} introduces some definitions and notation used throughout the paper and briefly recalls the principle of the foundational algorithm proposed in \cite{AlamirMazen2025Aafs}. Section \ref{sec-algo} explains the features space partitioning principle and provides the details of the algorithm leading to the implementation of aggregation-based prediction \eqref{newexplicit}. In Section \ref{sec-results}, the algorithm is validated on six-axis industrial manipulator robot\footnote{This is a more complicated use-case than the already involved four-axis robot used in  previous work regarding implicit relationships identification \cite{AlamirMazen2025Aafs}.} through an extensive comparison with a set of possible DNN architectures regarding precision and computation times. The additional experimental investigation of the generalization capacity to unseen contexts for a four-axis manipulator robot is also provided and discussed together with a comparison to the results provided by the best DNN architecture.
\section{Definitions and Notation}\label{sec-defnot}
\subsection{Datasets, features \& labels}
\noindent We consider a set of $n$ sensors whose values are considered as features\footnote{Here we use Machine Learning wording according to which one seeks a function $F$ such that the \textit{label} $y$ can be approximated as a function of the \textit{features} $x$ via $y\approx F(x)$.}:
$$x:= \begin{bmatrix} 
x^{(1)}\cr \vdots\cr x^{(n)}
\end{bmatrix}\in \mathbb R^n
$$
together with a sensor $y$ used as targeted label, namely to be expressed as a function of $x$. The value provided by these sensors at a sampling acquisition instant indexed by $t$ are denoted by $x_t\in \mathbb R^n$, $x^{(i)}_t\in \mathbb R$ and $y_t\in \mathbb R$. 
\e 
A dataset $\mathbb D$ is obtained by collecting time-series of $x$ and $y$ at a set of instants $\mathcal T$:
\begin{equation}
\mathbb D := \Bigl\{(x_t,y_t)\Bigr\}_{t\in \mathcal T}\in \Bigl[\mathbb R^n\times \mathbb R\Bigr]^{c_\mathcal T} \label{defdeD}
\end{equation}
where for any discrete set $\mathbb D$, $c_\mathbb{D}$ stands for the cardinality of $\mathbb D$, namely, the number of rows or samples. 
\e 
Given a dataset $\mathbb D$, the projection on the \textit{features} [resp. label] space are denoted by $\mathbb D\vert_x$ [resp. $\mathbb D\vert_y$],  namely:
\begin{equation}
\mathbb D\vert_x := \Bigl\{x_t\Bigr\}_{t\in \mathcal T}\quad;\quad \mathbb D\vert_y := \Bigl\{y_t\Bigr\}_{t\in \mathcal T}\label{Dverts}
\end{equation}
The derivation of the explicit relationships relies on the partitioning of the space of features into sub-regions. The notation associated to this task is discussed in the next section. 
\subsection{Partitioning triplets}\label{sec-partitioners}
\noindent Consider a dataset $\mathbb D$ and its projection on the space of features $\mathbb D\vert_x$. The following concept is extensively used in the sequel:
\begin{center}
\tikz{
\node[rounded corners, fill=Black!5, inner sep=3mm]{
\begin{minipage}{0.45\textwidth}
\begin{definition}
A partitioning triplet $\xi:=(\pi,n_d,x_c)$ consists in the following items:
\begin{itemize}
\item A polynomial $\pi:\mathbb R^n\rightarrow \mathbb R$,
\item An integer\footnote{This represents the number of sub-regions.} $n_d\ge 2$,
\item A centre\footnote{The partitioning is based on the value of $\pi(x-x_c)$.} $x_c\in \mathbb R^n$.
\end{itemize}
These items are also sometimes referred to using the following notation $\xi_\pi$, $\xi_{n_d}$ and $\xi_{x_c}$.
\end{definition}
\end{minipage}
};
}
\end{center}
A partitioning triplet enables to partition the space of features following the steps described below where $\pi$, $n_d$ and $x_c$ associated to the triplet are supposed to be fixed.
\e 
Consider the set of scalar values defined by:
\begin{equation}
\pi\bigl(\mathbb D\vert_x-x_c\bigr):=\Bigl\{\pi(x-x_c)\ \vert \ x\in \mathbb D\vert_x\Bigr\} \label{defdesetofz}
\end{equation}
together with the associated set of quantiles extended left and right by $-\infty$ and $+\infty$:
\begin{equation*}
\mathcal Q:=\Bigl\{-\infty,p_{\lfloor 100/n_d\rfloor},\dots,p_{\lfloor 100*(n_d-1)/n_d\rfloor},+\infty\Bigr\}
\end{equation*}
where $p_q$ is the $q$-th quantile of the set of numbers $\pi(\mathbb D\vert_x-x_c)$.
\e For instance, for $n_d=3$, the set $\mathcal Q$ is defined by the sequence $\{-\infty,p_{33\%},p_{66\%}+\infty\}$
leading to the following three intervals of values
$\Bigl\{(-\infty,p_{33\%}]\ ,\ (p_{33\%}, p_{66\%}] \  \text{and}\   (p_{66\%},+\infty)\Bigr\}$.
\e 
More generally, a partitioning triplet $\xi=(\pi, n_d, x_c)$ enables to define $n_d$ intervals  covering all possible values of $\pi(x-x_c)$ and hence inducing a partitioning of the space of features according the index of the interval to which $\pi(x-x_c)$ belongs. The resulting partitioning map is denoted by:
\begin{equation}
\sigma_\xi(x)\in \{1,\dots,n_d\} \label{defdesigmax}
\end{equation}
and the associated $n_d$ regions are denoted by:
\begin{equation}
\mathcal R^{(i)}_\xi:= \Bigl\{x\in \mathbb R^n\ \vert\ \sigma_\xi(x)=i\Bigr\} \quad \forall i\in \{1,\dots,n_d\}\label{defdemathcalR}
\end{equation}
\subsection{Implicit piece-wise representation}\label{sec-implicit}
\e As it is mentioned earlier, the contribution of the present paper leverages a recently proposed algorithm \cite{AlamirMazen2025Aafs} that solves a specific implicit residual generation problem. For the sake of completeness, let us briefly recall some of the notation and ideas associated to the previous contribution. Assume that a working dataset $\mathbb D$ is available gathering instances of a vector of features $x$ and the associated values of some targeted label denoted by $y$. 
\e The problem of parsimoniously identifying an implicit multi-variate polynomial residual expressing relationships between $x$ and $y$ can be stated as follows:
\begin{center}
\begin{tikzpicture}
\node[rounded corners, fill=Black!5, inner xsep=4mm, inner ysep=6mm](O){
\begin{minipage}{0.4\textwidth}
Find a set of $n_m$ sparse polynomials 
\begin{equation}
P_\kappa:\mathbb R^n\rightarrow \mathbb R\qquad \kappa\in \{1,\dots, n_m\} \label{setofpol}
\end{equation} where
\begin{equation}
P_\kappa(x):=\sum_{j=1}^{n_c^{(\kappa)}}\alpha_j^{(\kappa)}\underbrace{\left[\prod_{i=1}^n [x^{(i)}]^{p_{ij}^{(\kappa)}}\right]}_{\phi_j^{(\kappa)}(x)} \label{defdePkappa}
\end{equation}
s.t. the following implicit residual:
\begin{equation}
e(x,y) := \min_{\kappa=1}^{n_m}\Bigl\vert y - P_\kappa(x)\Bigr\vert \label{defdeimplicitebis}
\end{equation}
is parsimoniously minimized over the training set $\mathbb D$. 
\end{minipage}
};
\node[rounded corners, fill=white, inner sep=1mm, draw=black] at(O.north){\footnotesize The $\mathbb D$-Implicit piece-wise polynomial ML problem};
\end{tikzpicture}
\end{center}
In \cite{AlamirMazen2025Aafs}, the precise meaning of the minimization involved in \eqref{defdeimplicitebis} is rigorously defined. Shortly speaking, an initial threshold \texttt{th} is defined on the regression error, and several attempts are made to meet this error in a parsimonious way, beyond which the threshold is progressively increased until the training dataset is covered. 
% As in any algorithm, there is a set of hyper-parameters to be chosen a priori, the main such hyper-parameters involved in the algorithm proposed by \cite{AlamirMazen2025Aafs} that might be referred to in the remainder of the present paper are: 
% \begin{itemize}
% \item[$\checkmark$] The degree of the polynomials $P_\kappa$.
% \item[$\checkmark$] The precision threshold, denoted by $\texttt{th}\in (0,1)$, that sets a condition to accept a  polynomial as one of the $n_m$ polynomials involved in \eqref{defdeimplicitebis}. 
% \end{itemize}

\section{Identification of explicit piece-wise relationships}\label{sec-algo}
\noindent The first idea that comes to mind when searching for an explicit model is to keep track of the optimal indices, say $\kappa^\star(x,y)\in \{1,\dots, n_m\}$ for $(x,y)\in \mathbb D$, upon running the algorithm proposed in \cite{AlamirMazen2025Aafs} which solves \eqref{defdeimplicitebis}, for the training dataset $\mathbb D$, and to use the resulting dataset: 
\begin{equation}
\Bigl\{x,\kappa^\star(x,y)\Bigr\}_{(x,y)\in \mathbb D} \label{defdetraining}
\end{equation}
in which $x$ plays the role of features vector while the integer $\kappa^\star\in \{1,\dots,n_m\}$ plays the role of label (class index) to train an $n_m$-class ML-classifier which might be able to provide a features-dependent polynomial's index, namely:
\begin{equation}
 \texttt{cl}:x\rightarrow \{1,\dots,n_m\}\label{defdeclpasmarcher}
\end{equation}
Indeed, if the task of building such a classifier were successful, the explicit formulae one is looking for would simply bowl down to:
\begin{equation}
\hat y:= P_{\texttt{cl}(x)}(x) \label{defdefauxexplicit}
\end{equation}
Unfortunately, the author's experience on different datasets showed that this roadmap is quite often unsuccessful in that the precision of the resulting classifier is not good enough for the resulting prediction to be acceptable\footnote{As a matter of fact, the precision of the resulting piece-wise polynomial explicit relationship is very often \textit{worse} that a single explicit polynomial relationships. This can be explained given the intimate details of the algorithm of \cite{AlamirMazen2025Aafs} but such an explanation would go beyond the scope of the present contribution for which only the failure of this first attempt does really matter as it justifies the new heuristic presented in this contribution.}.
\e 
This is precisely the reason why the methodology proposed in the present section borrows a different path that is based on an a priori set of explicit partitions of the space of features that is not derived from to the values of indices $\kappa^\star$ encountered during the execution of the implicit form computation's algorithm. 
\e 
More precisely and anticipating the presentation that follows, it will be shown that each partitioning triplet $\xi:=(\pi,n_d,x_c)$ as defined in Section \ref{sec-partitioners} provides a single predictor (voter) amongst a set of  $n_V$ voters involved in the definition \eqref{newexplicit}. So let us first explain how such a partitioning triplet $\xi$ leads to an explicit piece-wise polynomial predictor (voter). This is the aim of the next section. Once this is done, the rationale according to which several voters are iteratively generated is explained in Section \ref{sec-severalVoters}.
\subsection{A single $\xi$-associated predictor}
\noindent This is where the concept of partitioning triplet introduced in Section  \ref{sec-partitioners} becomes helpful. Indeed, each candidate partitioning triplet $\xi=(\pi,n_d,x_c)$ induces a partitioning of the space of features into $n_d$ regions $\bigl\{\mathcal R^{(i)}_\xi\bigr\}_{i=1}^{n_d}$ defined by \eqref{defdemathcalR}. Now, it is possible to associate to each of these  regions the index of the \textit{best}  polynomial among those involved in the implicit solution, namely $P_\kappa, \kappa=1,\dots,n_m$. More precisely: 
\begin{equation}
\forall i\in \{1,\dots,n_d\},\ 
\kappa_\xi(i):=\texttt{arg}\min_{\kappa\in \{1,\dots,n_m\}}\sum_{x_t\in \mathcal R^{(i)}_\xi}\Bigl\vert P_\kappa(x_t)-y_t\Bigr\vert \label{defdekappastardots}
\end{equation}
Namely, for each region $\mathcal R_\xi^{(i)}$, the best polynomial is the one that minimizes the Mean-Absolute-Error MAE\footnote{Obviously the Mean-Squared-Error (MSE) can be used as well.}  over the instances lying inside the region $\mathcal R_\xi^{(i)}$. This enables to define an explicit relationship for each partitioning triplet $\xi$ as follows:
\begin{center}
\begin{tikzpicture}
\node[rounded corners, fill=Black!5, inner sep=3mm](M){
\begin{minipage}{0.45\textwidth}
Given\\
\begin{itemize}
\item[$\checkmark$] A set of polynomials $P_\kappa$, $\kappa=1,\dots,n_m$ involved in an implicit model;
\item[$\checkmark$] A partitioning  triplet $\xi:=(\pi,n_d,x_c)$ leading to the set of regions $\mathcal R^{(i)}_\xi$, $i=1,\dots,n_d$,
\end{itemize}
\ \\ an explicit relationship is defined that associates to each vector of features $x\in R^{(i)}_\xi$ the prediction given by $P_{\kappa_\xi(i)}(x)$ where the index $\kappa^{(i)}_\xi$ is defined by \eqref{defdekappastardots}. 
\end{minipage}
};
\node[rounded corners, fill=white, draw=Blue] at(M.north){\footnotesize $\xi$-associated explicit model};
\end{tikzpicture}
\end{center}
However, using the fact that the index $i$ of the region to which $x$ belongs is precisely $\sigma_\xi(x)$ [see \eqref{defdesigmax}], the explicit relationship induced by the partitioning triplet $\xi$ can be condensed into:
\begin{equation}
\mathcal P_\xi(x):= P_{\kappa_\xi(\sigma_\xi(x))}(x)\label{defdecalPx}
\end{equation}
hence, defining the $\kappa^\star$ map involved in \eqref{newexplicit} by $\kappa^\star:=\kappa_\xi\circ\sigma_\xi$.
\e 
As will be further explained, the above explicit predictor is based on a process that depends on a randomly generated polynomial $\pi$, so the resulting fit might obviously be far from being optimal. The idea is thus to create a set of different explicit representations providing a set of predictions that are then aggregated to provide the final explicit prediction. This is detailed in the next section. 
\subsection{A multiple voters-based prediction}\label{sec-severalVoters}
\noindent The main idea is to start with an initial partitioning triplet $\xi^{(1)}$ with centre $x_c^{(1)}=0$ and a randomly generated polynomial\footnote{The random generation is done for a priori given degree and number of active monomials.} $\pi^{(1)}$. The degree, say $\texttt{deg}_\text{part}\in \mathbb N$ of the partitioning polynomial is fixed as well as the number $n_d$ of partitioning intervals. These two parameters are used identically for all the  forthcoming partitioning triplets (voters). 
\e The resulting explicit piece-wise polynomial $\mathcal P_{\xi^{(1)}}$ defined by \eqref{defdecalPx} induces an initial precision $\epsilon^{(1)}$, computed as shown hereafter using \eqref{defdeepsilon}. This precision is then used as basis for the possible acceptance of the next voter, namely $\xi^{(2)}$. More precisely, when another candidate partitioning triplet $\xi^{(2)}$ is examined, it is accepted only if the resulting new prediction law:
\begin{equation*}
\mathcal P_\texttt{Voters}(x):=\dfrac{1}{2}\sum_{\ell=1}^2\mathcal P_{\xi^{(\ell)}}(x)\  \vert\  \texttt{Voters}:=\{\xi^{(\ell)}\}_{\ell=1}^2
\end{equation*}
provides a precision that is lower than $\gamma_c\times \epsilon^{(1)}$ for a predefined contraction parameter $\gamma_c\in (0,1)$. 
\e 
More generally, when a current number $j-1$ of voters has been accepted, adding a $j$-th one is conditioned by the prediction:
\begin{equation}
\mathcal P_\texttt{Voters}(x):=\dfrac{1}{j}\sum_{\ell=1}^j\mathcal P_{\xi^{(\ell)}}(x)\  \vert\  \texttt{Voters}:=\{\xi^{(\ell)}\}_{\ell=1}^j\label{defdesum}
\end{equation}
providing a precision $\epsilon^{(j)}$ such that 
\begin{equation}
\epsilon^{(j)}\le \gamma_c\times \epsilon^{(j-1)} \label{defdegammmac}
\end{equation}
The process stops when the number of trials $n_V^{max}$ is reached. For a given set of voters, the precision is computed as follows:
\begin{equation}
\epsilon(\texttt{Voters}\vert \mathbb D) := \mu\Biggl[\Bigl\{\mathcal P_\texttt{Voters}(x)-y\Bigr\}_{(x,y)\in \mathbb D}\Biggr] \label{defdeepsilon}
\end{equation}
where $\mu$ is some measure of the regression errors computed on the training data. 
\e 
When $j$ voters are already adopted, the new candidate partitioning triplet is generated as follows: The training samples are first reordered according to the regression error. Based on this reordering, the centre $x_c^{(j)}$ is randomly chosen among the top $\eta_\text{top}=5\%$ higher errors samples. As for the polynomial $\pi^{(j)}$, it is randomly generated using the pre-defined degree $\texttt{deg}_\text{part}$ and the pre-defined number $n_\texttt{Modes}^\texttt{part}$ of active monomials. This can be summarized in Algorithm \ref{algo}. 
\begin{algorithm}
\footnotesize
\caption{Explicit piece-wise polynomial design}\label{algo}
\begin{algorithmic}[1]  % The [1] adds line numbering
\State \textbf{Input parameters}: \\ 
$\{P_\kappa\}_{\kappa=1}^{n_m}$, $n_V^\text{max}$, $\gamma_c$, $\texttt{deg}_\text{part}$, $\eta_\text{top}$, $n_d$, $n_\texttt{Modes}^\texttt{part}=20$
\State \textbf{Initialization}: $\xi^{(1)}\leftarrow (\pi^{(1)},n_d,0)$, $\texttt{Voters}\leftarrow \{\xi^{(1)}\}$, \\ $\texttt{n}_v\leftarrow 1$, $\texttt{iter}\leftarrow 1$, $\epsilon^{(1)}\leftarrow \epsilon(\texttt{Voters}\vert\mathbb D)$. \\ $n_\text{top}\leftarrow \lfloor \eta_\text{top}\times c_\mathbb{D}\rfloor$ \Comment{{\scriptsize\color{Gray} Number of top error rows}}
    \While{$\texttt{iter}\le n_V^\text{max}$}
        \State $e\leftarrow \vert \mathcal P_\texttt{Voters}(\cdot)-y(\cdot)\vert$ \Comment{{\scriptsize\color{Gray} Compute errors}}
        \State $X_\text{ord}\leftarrow \texttt{sort}(\mathbb D\vert_x\ \text{according to increasing}\  e)$\Comment{{\scriptsize\color{Gray}Sort features rows}}.
        \State $x_c\leftarrow \texttt{random}(X_\text{ord}[:-n_\text{top}])$\Comment{{\scriptsize\color{Gray} Select a random row}}.
        \State $\pi\leftarrow$ Random polynomial of degree $\texttt{deg}_\text{part}$.
        \State $\xi\leftarrow (\pi,n_d,x_c)$ \Comment{{\scriptsize\color{Gray}Candidate partitioning triplet}}.
        \State \texttt{Voters\_cand} $\leftarrow$ \texttt{Voters} $\cup \{\xi\}$ \Comment{{\scriptsize\color{Gray}Candidate set of voters}}
        \State $\epsilon_\text{cand}\leftarrow \epsilon(\texttt{Voters\_cand}\vert \mathbb D)$ \Comment{{\scriptsize\color{Gray}The candidate error}}
        \If{$\epsilon_\text{cand}\le \gamma_c\times \epsilon^{(n_v)}$}
        \State $\texttt{Voters}\leftarrow \texttt{Voters\_cand}$ \Comment{{\scriptsize\color{Gray} Adopt the set of Voters}}
        \State $n_v\leftarrow n_v+1$ \Comment{{\scriptsize\color{Gray} Update the number of voters}}
        \State $\epsilon^{(n_v)}\leftarrow \epsilon_\text{cand}$ \Comment{{\scriptsize\color{Gray} Update the new precision}}
        \EndIf
        \State $\texttt{iter}\leftarrow \texttt{iter}+1$
    \EndWhile
    \State \textbf{return} $a$
\end{algorithmic}
\end{algorithm}
\e 
The following comments help understanding the steps of the algorithm. 
\e 
$\checkmark$ The input parameters of the algorithm are the ones that have been already discussed, namely: The set of polynomials involved in the implicit piece-wise polynomial solution $\{P_\kappa\}_{\kappa=1}^{n_m}$, the maximum number $n_V^\text{max}$ of voters, the contraction rate $\gamma_c\in (0,1)$ involved in \eqref{defdegammmac}, the degree $\texttt{deg}_\text{part}$ and the number $n_\texttt{Modes}^\texttt{part}$ of active monomials used in the partition polynomials $\pi$ involved in the candidate partition triplets $\xi^{(\ell)}$ and finally, the top ratio $\eta_\text{top}\in (0,1)$ used to define the highest error features vector inside which the future value of the centre $x_c$ is to be randomly sampled. 
\e 
$\checkmark$ The set of voters, denoted by $\texttt{Voters}$ is represented by the set of partitioning triplets that are designed upon executing the algorithm. It is initialized in Step 3 by randomly generating a polynomial $\pi^{(1)}$ while setting $n_d^{(1)}$ and $x_c^{(1)}$ to respectively $n_d$ and $0\in \mathbb R^{n_x}$ (Step 3). The current eligibility precision error $\epsilon^{(1)}$ is computed according to \eqref{defdeepsilon} in Step 4. Based on $\eta_\text{top}$ the number of top error rows $n_\text{top}$ in the training data is computed (Step 5). 
\e 
$\checkmark$ Steps 6-11 enable to generate a new \textit{candidate} partitioning triplet. This is done by randomly selecting a new candidate centre $x_c$ among the $n_\text{top}$ highest errors rows (samples) in the training data, the error is computed in Step 7. This enables to reorder the set of rows (samples) in the training dataset (Step 8). A random selection is performed among the $n_\text{top}$ highest error samples (Step 9). In order to complete the items of a partitioning triplet, a random polynomial $\pi$ is sampled (Step 10) and the candidate partitioning triplet $\xi$ is defined (Step 11).
\e 
$\checkmark$ The new candidate partitioning triplet is used to create a \textit{candidate} set of voters (Step 12) for which the associated candidate error $\epsilon_\text{cand}$ is computed (Step 13). 
\e 
$\checkmark$ Steps 14-18 check whether the required contraction rate ($\gamma_c$) on the precision level is satisfied [see \eqref{defdegammmac}] in which case, the set of voters, the number of voters and the new precision threshold are updated accordingly in Steps 15-17. Otherwise, a new iteration is fired leading to a new hopefully successful partitioning triplet. 
\e 
$\checkmark$ The algorithm stops when the number of allowable trials $n_V^\text{max}$ is reached (Step 6).

\begin{rem}[Two sets of polynomials]
Notice that the previously described solution involves two sets of polynomials which play two different roles in the resulting explicit model:
\begin{itemize}
    \item[$\checkmark$] The set of polynomials denoted previously by $P_\kappa$, $\kappa\in \{1,\dots,n_m\}$ which represents an input argument for the algorithm proposed in the paper. These polynomials are obtained using the algorithm of \cite{AlamirMazen2025Aafs} and might be used to compute very tight implicit residuals of the form \eqref{defdeimplicite} once a pair $(x,y)$ is available.
    \item[$\checkmark$] The set of polynomials $\pi^{(\ell)}$, $\ell \in \{1,\dots,n_v\}$ which are used, each for a specific voter, to partition the features space in a different way and hence to derive $n_v$ different explicit relationships using the previous set of polynomials. 
\end{itemize}
\end{rem}

\section{Validation on industrial data}\label{sec-results}
\e The framework proposed in the previous sections is evaluated on two use-cases concerning respectively a 6-axis (Figure \ref{fig-staubliTx290}) and a 4-axis (Figure \ref{fig-robot4axes}) manipulator robots. While the former use-case focuses on the ability of the algorithm to scale in the presence of a decently high number of features ($\texttt{dim}(x)=18$), the latter has been used to collect two datasets corresponding to different contexts of use where the ability of the models to generalize to unseen contexts can be more aggressively challenged\footnote{As the need for this second use-case appeared in a later stage of the research program, the first robot was no more available for a new round of multiple-context focused measurement collection.}.
\subsection{Use-case 1: Scalability, sparsity, performance and computation time}
\begin{figure}[h]
\begin{center}
\includegraphics[width=0.2\textwidth]{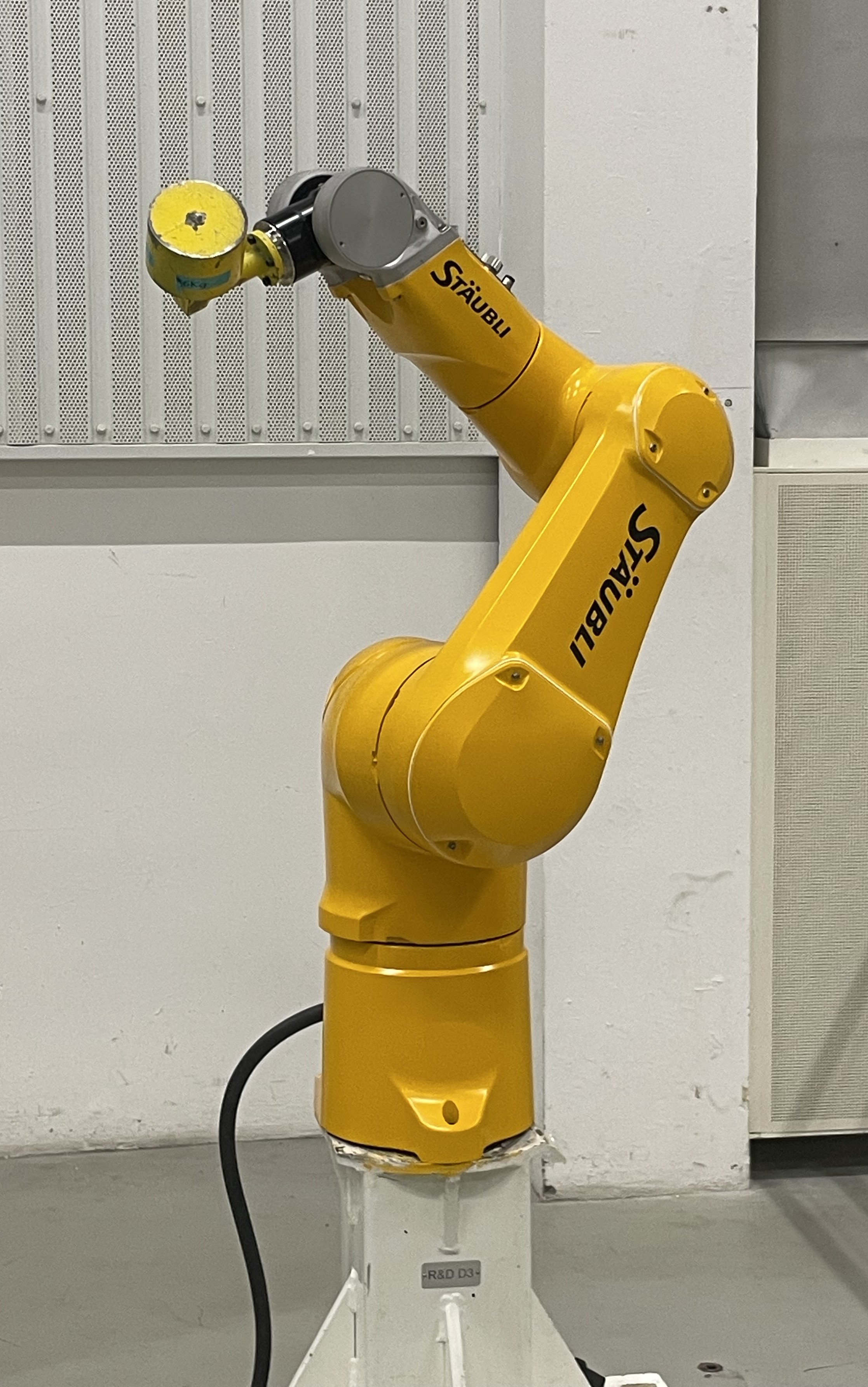} 
\end{center}   
\caption{The 6-axis \textsc{Staubli TX2-90} used in the present paper to validate the proposed algorithms and heuristics.}\label{fig-staubliTx290}
\end{figure}

\noindent In this first use-case, the 6-axis robot depicted in Figure \ref{fig-staubliTx290} is used. The dataset, the algorithm's hyperparameters setting, the performance indicators, the used hardware as well as the alternative DNNs architectures to which comparison is made are successively presented. Some of the information are shared with the second examples described in Section \ref{sec-secondusecase}.
\subsubsection{The Dataset}
\noindent The whole dataset includes 381 different random trajectories of a 6-axis manipulator robot (see Figure \ref{fig-staubliTx290}). The mean duration of a single trajectory is 51 sec with a standard deviation of 4.1 sec. The measurements are acquired at a frequency of 250 Hz leading to a total dataset including \textbf{4,858,087} rows (samples). 
\e The set of measurements comprises the kinematic variables: $q\in \mathbb R^6$, $\dot q\in \mathbb R^6$, $\ddot q\in \mathbb R^6$ used to define the vector of features $x:=(q,\dot q, \ddot q)\in \mathbb R^{18}$. Six labels are eligible which are the torques $\tau_i$, $i=1,\dots,6$ applied to the robot joints. Notice that each of the six possibilities of the label choice leads to a different instance of the identification problem $\tau_i=F_i(x)$, $i=1,\dots,6$  which all share the same vector of features $x\in \mathbb R^{18}$. 
\e 
\tikz{
\node[rounded corners, fill=black!5, inner ysep=5mm, inner xsep=2mm, draw=Gray](T){
\begin{minipage}{0.45\textwidth}
Only one over 10 samples in the first 5\% part of the dataset (no shuffle) is used for training leading to a total number of training samples of \textbf{24,290}. The test dataset involves \textbf{4,615,183} samples.
\end{minipage}
};
}
The above splitting choice is intended to challenge the ability of the algorithm to provide a relevant prediction model despite a quite small number of samples and the decently high dimension of the features space. Notice however that all the trajectories, while quite different, still belong to the same \textit{family} (i.e. joint trajectories) and show the same amplitudes of excursion which makes the harmful consequences of overfitting less easy to show hence justifying the second use-case examined later in Section \ref{sec-secondusecase}. 
\subsubsection{Algorithm's hyperparameters setting}\label{sec:hyper1}
\noindent For all the problem's instances, the degrees of the polynomials $P_\kappa$ used in the piece-wise polynomial identification as well as for those, denoted by $\pi^{(\ell)}$ and used in the definition of the partition triplets is set to $3$. The contraction parameter $\gamma_c$ is set to 0.999. The parameter $\eta_\text{top}=5\%$ is used in the selection of the new centres $x_c$. The maximum number of voters is set to $n_V^\text{max}=20$. The number of intervals $n_d=50$ is used for the definition of the features space partitioning triplets. The number of random monomials that are used to define the polynomials $\xi_\pi$ associated to a triplet $\xi$ is chosen to be $n_\texttt{Modes}^\texttt{part}=20$.
\subsubsection{Performance indicators}
\noindent The quality of the models is measured through the normalized Mean Absolute Error (nMAE) and the normalized Mean Squared Error (nMSE) which are defined as follows: 
\begin{subequations}\label{perfindic}
\begin{align}
\texttt{nMAE} &:= \dfrac{\texttt{MAE}(y_\text{pred}(\cdot)-y_\text{true}(\cdot))}{\texttt{MAE}(y_\text{true}(\cdot))}\label{nMAE}\\
\texttt{nMSE} &:= \dfrac{\texttt{MSE}(y_\text{pred}(\cdot)-y_\text{true}(\cdot))}{\texttt{MSE}(y_\text{true}(\cdot))}\label{nMSE}
\end{align}
\end{subequations}
A comparison is provided with different structures of DNNs for which it is a common practice to monitor the ratio of the residual computed on the \textit{validation} dataset\footnote{The validation dataset is used in order to provide an early stopping criterion which enables to partially avoid overfitting during the DNN learning step.} to the residual obtained on the training dataset. This ratio is commonly referred to as the \textit{generalization gap}.
\e Now since the proposed algorithm does not involve splitting the learning data into training and validation (no gradient descent is used and the risk of overfitting is contained by the very definition of the parsimonious regression), we shall use a posteriori measure of the \textit{generalization gap}, which involves the test dataset instead of the validation dataset, as the former is available for both DNNs and the proposed framework. 
\e More precisely, we consider the so called \textit{centred generalization gap} defined by:
\begin{equation}
\alpha_c:= \Biggl\vert \dfrac{\texttt{nMAE}\vert_\text{test}}{\texttt{nMAE}\vert_\text{train}}-1\Biggr\vert \label{defdealphac}
\end{equation}
The reason for which the absolute value is used stems from the fact that, for the proposed solution, it is not always true that the residual on the training data is smaller than the residual on the test data. The centred criterion is such that small values of $\alpha_c$ indicate a high  generalization capacity, while values that are far from zero mean that the fitting process induced an amount of overfitting making the use of the model on unseen contexts riskier. 
\subsubsection{Hardware for computation times evaluation}\label{sec-hardware}
\noindent For technical reasons, the computations of the DNNs-based solutions and those involving the proposed algorithm have been performed on two different machines. Nevertheless, it is possible to use the associated computation time to still get a fair comparison, considering that the difference is more than an order of magnitude, and that the longer computations are obtained on the faster machine/implementation. (see Table \ref{tab-cpus})
\e
More precisely, as far as the computation of the proposed algorithm (denoted hereafter by \texttt{xpwpol} for explicit piece-wise polynomial) is concerned, the computation has been performed on a \texttt{MacBook Pro} with Processor \texttt{Apple M3 Pro}, 18Go RAM using \texttt{python 3.11.8} using only standard CPU computation. 
\e As for the DNN-related investigation, the computation has been performed on Windows machine with \texttt{13th Gen Intel Core i7-13700K}, 3400 MHz, 16 cores processor using \texttt{Pytorch} 2.1.2, \texttt{CUDA} 12.1 and a \texttt{NVIDA RTX A2000} (6 Go) GPU.
\subsubsection{Alternative DNN models involved in the comparisons}
\noindent In the present use-case, three different DNN structures are tested and compared to the proposed algorithm, namely: standard Multi-Layer Perception (MLP),  Convolutional Neural Network (CNN) and a Transformer architecture. All of them have been implemented using the \texttt{Pytorch} python module \cite{paszke2019pytorch} on the hardware described in Section \ref{sec-hardware}. 
\e Upon fitting the DNN models, a standard \textit{early stopping} mechanism is used based on splitting the learning dataset into a training and validation subsets. 
\e 
Regarding the \textit{early stopping} condition commonly used when training DNN models in order to avoid overfitting, it is hereafter defined by the iteration achieving the minimum residual on the validation dataset. Notice that some recent results \cite{VILARESFERRO2023109,ahmad2022when} suggest to be more cautious by considering a threshold on the generalization gap defined by \eqref{defdealphac} beyond which the gradient iterations should be stopped even if the validation residual keep decreasing. The rational is that high values of the generalization gap $\alpha_c$ mean that the model already starts overfitting on the training data and hence might be dangerous to use on future unseen contexts. 
\e 
\tikz{
\node[rounded corners, fill=Gray!10, inner xsep=3mm, inner ysep=5mm, draw=Gray](U){
\begin{minipage}{0.45\textwidth}
It has been decided to adopt the most commonly used early stopping condition (that minimizes the validation loss) leading to optimistic evaluation of the residuals of the DNN models which systematically show obvious amount of overfitting as it can be witnessed by high values of the generalization gap $\alpha_c$ (see Table \ref{tab:alphaSmall}).
\end{minipage}
};
\node[rounded corners, fill=white, draw=Gray] at(U.north){\footnotesize Rather optimistic residual values for DNN models};
} 
\begin{figure}[h]
\begin{center}
\includegraphics[width=0.48\textwidth]{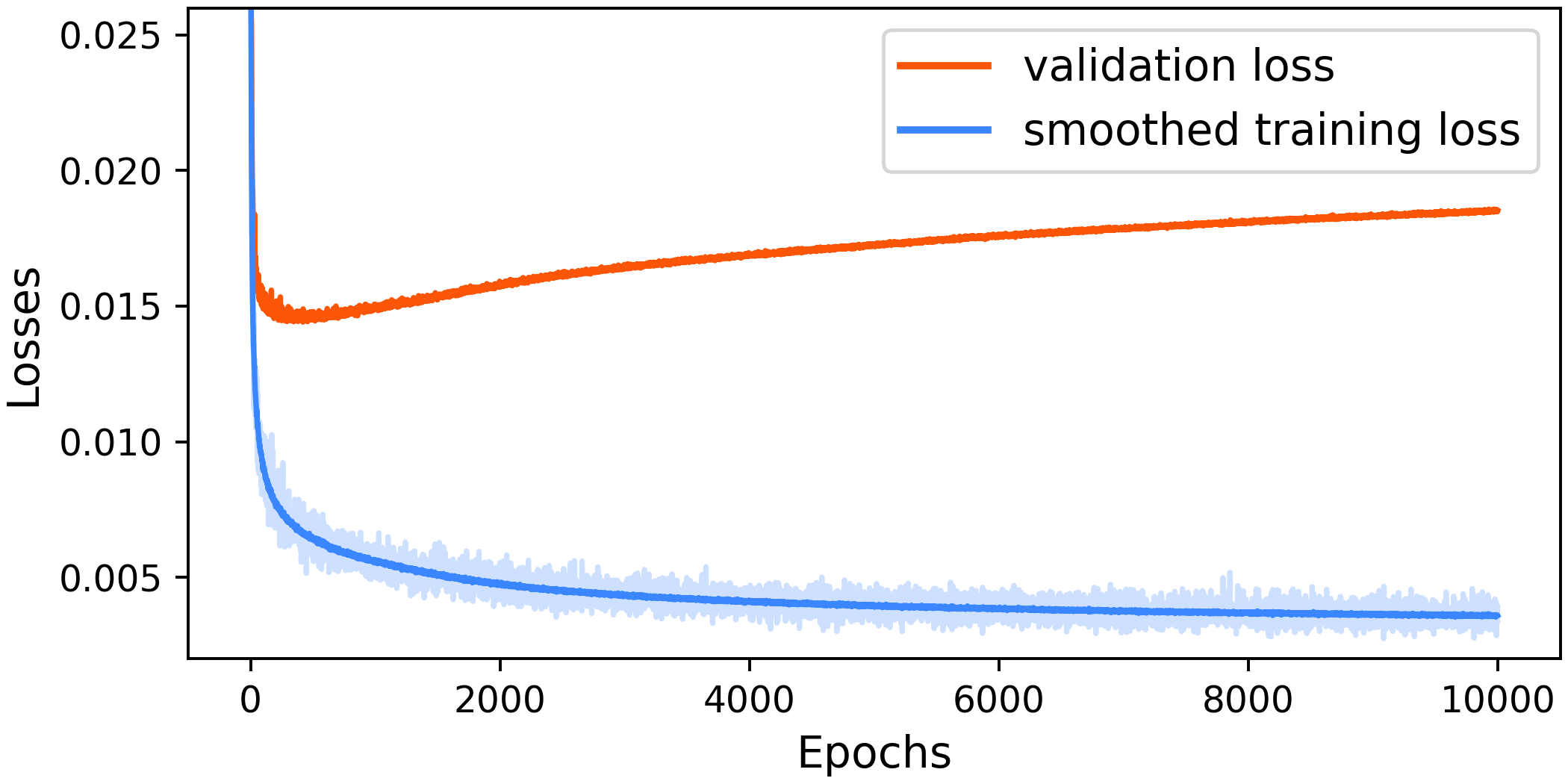} 
\end{center}
\caption{Use-case 1: Typical evolution of the losses behaviour on training and validation datasets during the learning of the DNN models.}\label{fig-loss}
\end{figure}
Figure \ref{fig-loss} shows typical evolutions of the loss functions computed on the training and validation datasets. In the specific case shown in Figure \ref{fig-loss}, when the iterations are stopped (at the iteration number that minimizes validation loss), there is already a significant gap between the training and the validation losses suggesting that a cautious learning would have been stopped quite earlier leading to significantly higher residuals than the ones reported in the forthcoming tables \ref{tab:ressmall} and \ref{tabreslarge}.
\e Now it can be argued that what do really matter are the performance results on the unseen test dataset. This is only partially true since, as it is highlighted above,  the trajectories used in all the datasets concerned by the present use-case, while rigorously different, still belong to the same family of generated trajectories. This reduces the risk of observing the bad consequences of overfitting on the test dataset.
\subsubsection{Results}
\begin{figure*}[t]
\begin{center}
\includegraphics[width=0.98\textwidth]{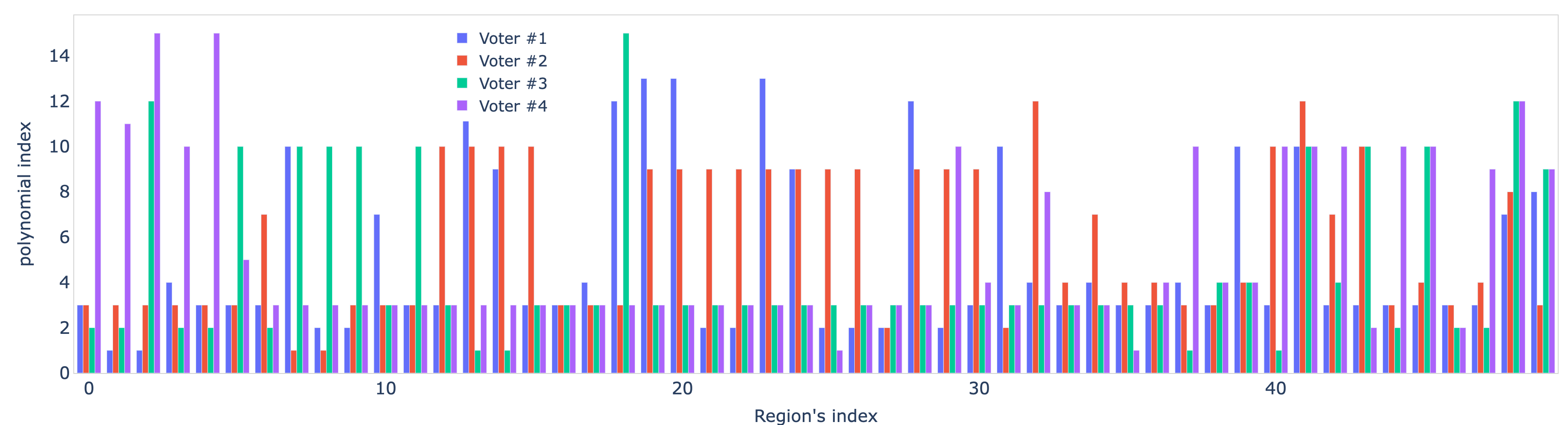}
\caption{Use-case 1: Definition of the partitioning maps $\kappa_{\xi^{(j)}}(i)\in \{1,\dots,15\}$ defined by \eqref{defdekappastardots} for the four voters included in the explicit piece-wise model for Axis 1. Notice that $j\in \{1,\dots,4\}$ refers to the voter index while $i\in \{1,\dots,n_d=50\}$ refers to the region's index. (see Table \ref{tab:nmnVoters} for a reminder regarding the properties of the model for Axis 1.) }\label{fig-kappaxi}
\end{center}
\end{figure*}

\noindent The execution of the proposed algorithms leads to six explicit piece-wise polynomial models showing the characteristics presented in Table \ref{tab:nmnVoters}. This table shows, for each identification problem corresponding to the choice of the label among the six torques applied to the robot's joints, the number $n_m$ of polynomials $P_\kappa$ involved in \eqref{defdeimplicite} which are obtained upon solving the implicit form and hence are input arguments for Algorithm \ref{algo}. The second row of Table \ref{tab:nmnVoters} shows the number of voters resulted from the execution of Algorithm \ref{algo} for each choice of the label. 
\begin{table}[h]
\footnotesize
\centering
\caption{Use-case 1: Number of polynomials and number of voters for the explicit piece-wise polynomial models}\label{tab:nmnVoters}
\begin{tabular}{ccccccc}
\toprule

\multirow{1}{*}{Axis}

& \multicolumn{1}{c}{\bf 1}
& \multicolumn{1}{c}{\bf 2} 
& \multicolumn{1}{c}{\bf 3} 
& \multicolumn{1}{c}{\bf 4}
& \multicolumn{1}{c}{\bf 5}
& \multicolumn{1}{c}{\bf 6}\\

%\cline{2-7}

% & nMAE & nMSE
% & nMAE & nMSE
% & nMAE & nMSE\\

\midrule

$n_m$ & 15 & 2 & 4 & 21 & 25 & 25\\
\texttt{nVoters} & 4  & 1 & 3 & 4 & 2 & 6 \\
\bottomrule
\end{tabular}
\end{table}
\e 
For the sake of illustration, Figure \ref{fig-kappaxi} shows the definition of the maps $\kappa_{\xi^{(j)}}(\cdot)$ for $j=1,\dots,4$ corresponding to the 4 voters involved in the explicit piece-wise polynomial corresponding to axis 1. Notice that the x-axis refers to the index of the region among the $n_d=50$ regions used in the model while the $y$-axis shows the index of the polynomial $P_\kappa$ used in each specific region. 
\e 
Tables \ref{tab:ressmall} and \ref{tab:alphaSmall} show the comparison with DNNs of \textit{moderate} size. More precisely, Table \ref{tab:ressmall} shows the \texttt{nMAE} and the \texttt{nMSE} for the three DNN structures with moderate complexity and the proposed explicit piece-wise polynomial structure denoted by \texttt{xpwpol}. Table \ref{tab:alphaSmall} shows the corresponding generalization gap $\alpha_c$ as defined in \eqref{defdealphac}. In both tables, each row corresponds to one of the six axes of the manipulator robot of Figure \ref{fig-staubliTx290} when used as targeted label in the identification problem. Table \ref{tab:complexity} shows the number of active coefficients\footnote{For polynomials, the number of active coefficients represents the number of selected monomials while for DNNs, it represents the number of \textit{trainable} coefficients as all of them are a priori non rigorously zero in a gradient-descent based training.} in all the fitted models. 
\e It can be observed that the quality of the prediction of the models resulted from the proposed framework on the test datasets are quite comparable to those provided by the DNNs for the \textbf{first three axes} while the residuals of the explicit piece-wise polynomials on the remaining axes is quite higher (while keeping the same order of magnitude). Notice however that the generalization gap $\alpha_c$ increases also, on these axes, for the DNN models as the difference in residuals increases. On the other hand, the generalization gap is remarkably small for the explicit piece-wise polynomial models. 
\e 
Tables \ref{tabreslarge} and \ref{tab:alphacLarge} show the same information as Tables \ref{tab:ressmall} and \ref{tab:alphaSmall} but using the large DNN structures (notice that as far as the \texttt{xpwpol} structure is concerned, the results are identical to those shown previously, and are reproduced for the reader's convenience). Here again, Table \ref{tab:complexity} compares the number of coefficients involved in the large DNN models to those used in the models identified by the proposed framework.
\e 
Here again, the same comments can be reproduced regarding the residual values and the associated generalization gaps. Indeed, the latter increases significantly confirming the diagnosis regarding the overfitting introduced in the DNN models which increases for the minor axes 4, 5 and 6.
\e 
For instance, Table \ref{tab:alphacLarge} shows that the residual of MLP models for axis 5 on training data is 3.31 times (which is $\alpha_c+1$ in this case) smaller than its value in the test data.  Similarly, the Transformer model shows on the same axis a training residual that is 2.44 times smaller than the test residual. On the contrary, the \texttt{xpwpol} model shows a generalization gap of 0.03 meaning that the training and the test residuals are almost identical while providing comparable performance at least for the first three axes. 
\e 
As mentioned before, Table \ref{tab:complexity}  compares the complexity of  the different architectures, expressed in terms of the number of active coefficients. The difference in complexity explains the reasons for the different behaviour of the generalization gap $\alpha_c$ as the \texttt{xpwpol} models maximum number of coefficients never exceed 700 (with only 66 for the model for axis 2) while the DNN models number of active coefficients (i.e. the so called \textit{trainable parameters}) ranges from ten thousands to almost 800,000 coefficients. 
\e 
Finally, Table \ref{tab-cpus} shows the computation times needed to achieve the models' fitting (see Section \ref{sec-hardware} for more details) which is expressed in seconds for the \texttt{xpwpol} models while it is never lower than 2 hours for the MLP and the Transformer and always exceed 30 min for the CNN\footnote{The reason is that the CNN starts quite early to show overfitting behavior and hence the early stopping condition is activated earlier than for the other DNNs.}. Notice that for DNNs the computation time is computed assuming that the iterations stops at the minimum of the validation residual. 
%-----------
\begin{table*}
\footnotesize
\centering
\caption{Use-case 1: Results on the test dataset: Normalized residuals (nMAE, nMSE) $\vert$ Small DNN models}\label{tab:ressmall}
\begin{tabular}{ccccc}
\toprule

\multirow{1}{*}{Axis}

& \multicolumn{1}{c}{\bf MLP (Small)}
& \multicolumn{1}{c}{\bf CNN (Small)} 
& \multicolumn{1}{c}{\bf Transformer (Small)} 
& \multicolumn{1}{c}{\bf xpwpol}\\

%\cline{2-7}

% & nMAE & nMSE
% & nMAE & nMSE
% & nMAE & nMSE\\

\midrule

1 & (0.23, 0.06) & (0.24, 0.06) & (0.22, 0.06) & (0.22, 0.06)\\
2 & (0.09, 0.01) & (0.10, 0.01) & (0.10, 0.01) & (0.12, 0.02)\\
3 & (0.18, 0.04) & (0.16, 0.04) & (0.15, 0.03) & (0.20, 0.05)\\
4 & (0.25, 0.08) & (0.25, 0.08) & (0.27, 0.09) & (0.37, 0.17)\\
5 & (0.43, 0.20) & (0.43, 0.19) & (0.44, 0.20) & (0.60, 0.33)\\
6 & (0.30, 0.12) & (0.34, 0.15) & (0.33, 0.14) & (0.52, 0.31)\\

\bottomrule
\end{tabular}
\end{table*}

\begin{table*}
\footnotesize
\centering
\caption{Use-case 1: Centred Generalization Gap $\alpha_c$ defined by \eqref{defdealphac} $\vert$ Small DNN models}\label{tab:alphaSmall}
\begin{tabular}{ccccc}
\toprule

\multirow{1}{*}{Axis}

& \multicolumn{1}{c}{\bf MLP (Small)}
& \multicolumn{1}{c}{\bf CNN (Small)} 
& \multicolumn{1}{c}{\bf Transformer (Small)} 
& \multicolumn{1}{c}{\bf xpwpol}\\

%\cline{2-7}

% & nMAE & nMSE
% & nMAE & nMSE
% & nMAE & nMSE\\

\midrule

1 & 0.21 & 0.26 & 0.16 & \textbf{0.00}\\
2 & 0.50 & 0.43 & 0.43 & \textbf{0.01}\\
3 & 0.64 & 0.33 & 0.36 & \textbf{0.02}\\
4 & 0.39 & 0.39 & 0.42 & \textbf{0.11}\\
5 & 0.79 & 0.59 & 0.57 & \textbf{0.03}\\
6 & 0.76 & 0.70 & 0.57 & \textbf{0.09}\\

\bottomrule
\end{tabular}
\end{table*}

%-----------
\begin{table*}
\footnotesize
\centering
\caption{Use-case 1:  Results on the test dataset: Normalized residuals (nMAE, nMSE) $\vert$ Large DNN models}\label{tabreslarge}
\begin{tabular}{ccccc}
\toprule

\multirow{1}{*}{Axis}

& \multicolumn{1}{c}{\bf MLP (Large)}
& \multicolumn{1}{c}{\bf CNN (Large)} 
& \multicolumn{1}{c}{\bf Transformer (Large)} 
& \multicolumn{1}{c}{\bf xpwpol}\\

\midrule

1 & (0.23, 0.06) & (0.22, 0.05) & (0.20, 0.05) & (0.22, 0.06)\\
2 & (0.11, 0.02) & (0.09, 0.01) & (0.09, 0.01) & (0.12, 0.02)\\
3 & (0.19, 0.05) & (0.15, 0.03) & (0.15, 0.03) & (0.20, 0.05)\\
4 & (0.26, 0.09) & (0.24, 0.07) & (0.22, 0.06) & (0.37, 0.17)\\
5 & (0.53, 0.30) & (0.40, 0.16) & (0.39, 0.17) & (0.60, 0.33)\\
6 & (0.37, 0.17) & (0.28, 0.11) & (0.31, 0.12) & (0.52, 0.31)\\

\bottomrule
\end{tabular}
\end{table*}

\begin{table*}
\footnotesize
\centering
\caption{Use-case 1:  Centred Generalization Gap $\alpha_c$ defined by \eqref{defdealphac} $\vert$ Large DNN models}\label{tab:alphacLarge}
\begin{tabular}{ccccc}
\toprule

\multirow{1}{*}{Axis}

& \multicolumn{1}{c}{\bf MLP (Large)}
& \multicolumn{1}{c}{\bf CNN (Large)} 
& \multicolumn{1}{c}{\bf Transformer (Large)} 
& \multicolumn{1}{c}{\bf xpwpol}\\

%\cline{2-7}

% & nMAE & nMSE
% & nMAE & nMSE
% & nMAE & nMSE\\

\midrule

1 & 0.64 & 0.57 & 0.33 & \textbf{0.00}\\
2 & 1.20 & 1.25 & 0.80 & \textbf{0.01}\\
3 & 1.71 & 1.50 & 1.14 & \textbf{0.02}\\
4 & 1.00 & 1.00 & 0.69 & \textbf{0.11}\\
5 & 2.31 & 1.86 & 1.44 & \textbf{0.03}\\
6 & 1.64 & 1.55 & 1.38 & \textbf{0.09}\\

\bottomrule
\end{tabular}
\end{table*}

%-----------------------

\begin{table*}
\footnotesize
\centering
\caption{Use-case 1: Number of active coefficients.}\label{tab:complexity}
\begin{tabular}{clllc}
\toprule

\multirow{1}{*}{Axis}

& \multicolumn{1}{c}{\bf MLP}
& \multicolumn{1}{c}{\bf CNN} 
& \multicolumn{1}{c}{\bf Transformer} 
& \multicolumn{1}{c}{\bf xpwpol}\\

\midrule

1 &  &   &   & 387 \\
2 &  &   &   & 66 \\
3 & Small: 14,086 & Small: 10,542 & Small: 11,190 & 183\\
4 & Large: 617,990 & Large: 618,598  & Large: 797,958  & 464\\
5 &  &   &   & 431\\
6 &  &   &   & 688\\

\bottomrule
\end{tabular}
\end{table*}

%-----------------------

\begin{table*}[h]
\centering
\footnotesize
\caption{Use-case 1: Computation times for model fitting.}\label{tab-cpus}
\begin{tabular}{ccccc}
\toprule

\multirow{1}{*}{Axis}

& \multicolumn{1}{c}{\bf MLP}
& \multicolumn{1}{c}{\bf CNN} 
& \multicolumn{1}{c}{\bf Transformer} 
& \multicolumn{1}{c}{\bf xpwpol}\\

%\cline{2-7}

% & nMAE & nMSE
% & nMAE & nMSE
% & nMAE & nMSE\\

\midrule

1 &  &  &  & 106 sec\\
2 &  & (\texttt{pytorch-GPU}) &   & 35 sec\\
3 &  &  &   & 28 sec\\
4 & >2h & $>30$min & > 2h & 142 sec\\
5 &  &  &   & 162 sec\\
6 & & \scriptsize See Section \ref{sec-hardware} &   & 173 sec\\

\bottomrule
\end{tabular}
\end{table*}

\subsection{Use-case 2: Generalization over unseen contexts}\label{sec-secondusecase}
\begin{figure}[h]
    \centering
    \includegraphics[width=0.25\linewidth]{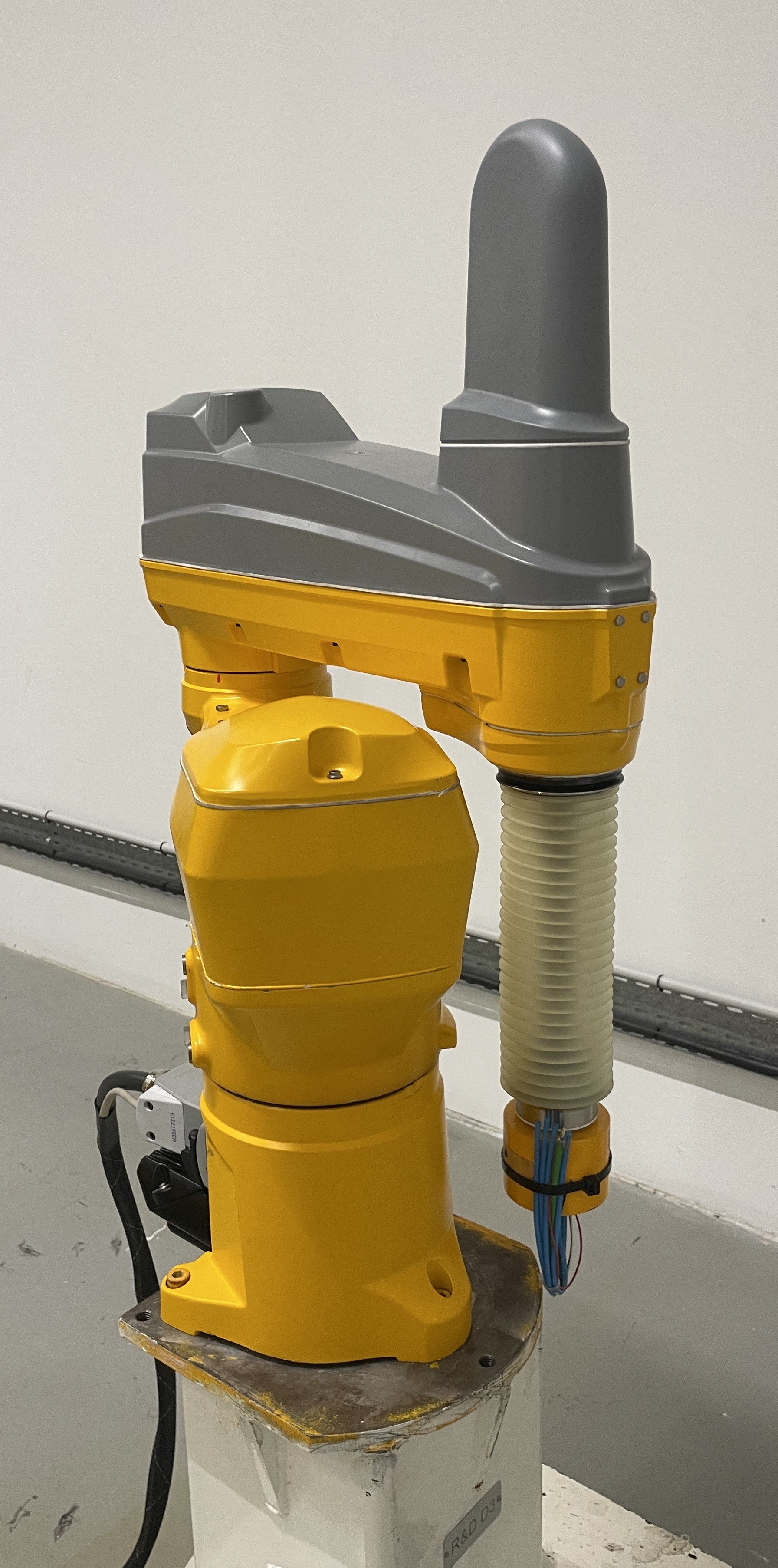}
    \caption{The 4-axes \textsc{Staubli TS0-80} used in the present paper to examine the generalization to unseen contexts.}\label{fig-robot4axes}
    \label{fig:robot4axes}
\end{figure}

\noindent In this second use-case, data are collected corresponding to two different contexts that differ by the amplitudes of the excursions used in the trajectories. Then the small amplitude trajectories are used in training (including the validation for DNN models) while the larger trajectories are used in the test. For availability reasons, this experiment is built using a 4-axis manipulator robot  such as the one depicted in Figure \ref{fig-robot4axes}. This enables to \textit{emulate} two different contexts of use so that the ability of models to generalize to unseen contexts can be challenged.
%635 200 835
\subsection{Dataset, metrics and alternative structure}
\begin{figure}
    \centering
    \includegraphics[width=\linewidth]{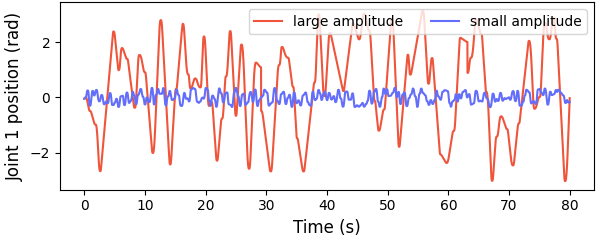}
    \caption{\color{Gray} Use-case 2: Typical difference in excursion between the trajectories used in the train/validation and \texttt{test-small} dataset (blue) and in the \texttt{test-large} dataset (red).}
    \label{fig:typicalDiff}
\end{figure}

\noindent The working dataset includes 835 different random trajectories. Among these trajectories, $635$ corresponds to small excursions while $200$ corresponds to large excursions. The number of trajectories for both subsets has been computed so that both contexts (small and large excursions) is represented by the same number of samples, namely $1,623,956$ leading to a total dataset involving \textbf{3,246,892} rows (samples). The measurements are acquired at a frequency of 250 Hz.
\e The set of measurements comprises the kinematic variables: $q\in \mathbb R^4$, $\dot q\in \mathbb R^4$, $\ddot q\in \mathbb R^4$ used to define the vector of features $x:=(q,\dot q, \ddot q)\in \mathbb R^{12}$. Four labels are eligible which are the torques $\tau_i$, $i=1,\dots,4$ applied to the axes.  
\e Figure \ref{fig:typicalDiff} shows a typical difference in terms of excursion between the trajectories used in the train/validation on the one hand and the test on the other hand. Obviously, in this experiment, the difference is intentionally exaggerated in order to underline the issue of generalization power. Naturally, smaller differences would have induced correspondingly smaller consequences on the generalization errors. 
\e Regarding the data splitting, Only one over 10 samples in the first 15\% of the small excursions dataset are used for training ($24,360$ samples), \textbf{the remaining of this small excursions-related  dataset is used for validation in order to monitor the early stopping condition, and for the model testing, referred to hereafter by \texttt{test-small}.} The final generalization test is done using the $1,622,936$ samples of the large excursions-related unseen contexts and the resuting test is referred to by \texttt{test-large}.
\e As for the DNN structure used in the comparison, given the results of Section \ref{sec-results}, only the large Transformer structure has been used as it stands out compared to the other ones.
\e 
The same performance indexes metrics discussed in the previous section are used hereafter. The discussion regarding the complexity and the computation times are not reproduced here as the order of magnitudes are similar to the ones depicted in Tables \ref{tab:complexity} and \ref{tab-cpus}.
\subsection{hyperparameters settings}
\noindent The hyperparameters of the proposed framework that have been used in the previous use-case (see Section \ref{sec:hyper1}) are identically used here except for the degrees of the polynomials $P_\kappa$ of the implicit solution that are taken all equal to $1$ (instead of 3 in the previous use-case). This emulates the situation where the solution's designer is aware of the risk of training/validation data being not fully representative of future operational use of the prediction model. 
\subsection{Results}
\noindent Table \ref{tab:nmnVotersUsecase2} shows the number $n_m$ of polynomials used in the solutions for the different axes, as well as the number of needed voters and the total number of active coefficients (total number of monomials involved in the expressions of the polynomials $P_\kappa$, $\kappa=1,\dots,n_m$). Remarkably enough, in this case, only a single partitioning triplet is used for all the models. 
\begin{table}[h]
\footnotesize
\centering
\caption{\color{Gray}  Use-case 2: Number of polynomials, number of voters and total number of active coefficients for the explicit piece-wise polynomial models. For the number of coefficients involved in the DNN models, see Table \ref{tab:complexity}.}\label{tab:nmnVotersUsecase2}
\begin{tabular}{lcccccc}
\toprule

\multirow{1}{*}{\bf Axis}

& \multicolumn{1}{c}{\bf 1}
& \multicolumn{1}{c}{\bf 2} 
& \multicolumn{1}{c}{\bf 3} 
& \multicolumn{1}{c}{\bf 4}\\

%\cline{2-7}

% & nMAE & nMSE
% & nMAE & nMSE
% & nMAE & nMSE\\

\midrule

$n_m$ & 4 & 5 & 5 & 8\\
\texttt{nVoters} & 1  & 1 & 1 & 1\\
\# coefficients & 273 & 104 & 194 & 282\\
\bottomrule
\end{tabular}
\end{table}

%-----------
\begin{table}[h]
\footnotesize
\centering
\caption{\color{Gray}  Use-case 2: Normalized residuals (nMAE, nMSE) $\vert$ \textbf{Test-small}}\label{tab:usecase2-perf-valid}
\begin{tabular}{ccc}
\toprule

\multirow{1}{*}{Axis}

& \multicolumn{1}{c}{\bf Transformer (test-small)}
& \multicolumn{1}{c}{\bf xpwpol (test-small)} \\

%\cline{2-7}

\midrule

1 & (0.08, 0.01) & (0.10, 0.01) \\
2 & (0.10, 0.01) & (0.12, 0.02) \\
3 & (0.13, 0.03) & (0.10, 0.01)\\
4 & (0.24, 0.10) & (0.20, 0.05)\\

\bottomrule
\end{tabular}
\end{table}

%-----------
\begin{table}[h]
\footnotesize
\centering
\caption{\color{Gray}  Use-case 2: Normalized residuals (nMAE, nMSE) $\vert$ \textbf{Test-large}}\label{tab:usecase2-perf-test}
\begin{tabular}{ccc}
\toprule

\multirow{1}{*}{Axis}

& \multicolumn{1}{c}{\bf Transformer (Test-large)} 
& \multicolumn{1}{c}{\bf xpwpol (Test-large)}\\

%\cline{2-7}

% & nMAE & nMSE
% & nMAE & nMSE
% & nMAE & nMSE\\

\midrule

1 & (1.38, 1.53) & (0.36, 0.13)\\
2 & (1.73, 2.57) & (0.28, 0.08)\\
3 & (0.77, 0.63) & (0.37, 0.14)\\
4 & (1.96, 3.79) & (2.33, 5.68)\\

\bottomrule
\end{tabular}
\end{table}
\e Tables \ref{tab:usecase2-perf-valid}  and \ref{tab:usecase2-perf-test} show the residual obtained respectively on the small excursions test dataset (the same context as in the training dataset) and on the large excursions (new context) dataset. 
\e Notice first of all that, Table \ref{tab:usecase2-perf-valid} suggests that despite the first order polynomials used in the derivation of the set of polynomials $P_\kappa$, $k=1,\dots,n_m$, involved in the original implicit representations (which are fed into Algorithm \ref{algo}), the explicit form obtained by Algorithm \ref{algo} which needs a single partitioning triplet (\texttt{nVoters}=1) enables to obtain an explicit \textbf{piece-wise affine} representations that show comparable residuals levels (if not smaller) on the small excursion test data (generalization to new trajectories within the same context) when compared to the Transformer's residuals and this for all the axis' models.
\e More interestingly, the results shown in Table \ref{tab:usecase2-perf-test} clearly show that the DNN model generalize very badly on truly unseen new contexts and this for all the axes. The parsimonious explicit  piece-wise affine models identified by Algorithm \ref{algo} show far more robust generalizations for the first three axes. On the contrary, for axis 4, generalization goes very badly as it is the case for the DNN model.
\section{Conclusion}\label{sec:conclusion}
\noindent A new heuristic is proposed in the present paper that enables to derive explicit parsimonious piece-wise polynomial representation expressing a label in terms of a vector of features that might involve a decently high number of components. The relevance of the proposed solution is shown through real-life experimental data collection on two industrial manipulator robots. Comparison with the state-of-the-art DNN alternatives is also provided. 
\e As stated by the famous \textit{no-free-lunch} theorem, the results show that there is no \textit{always winning} option and consequently, the proposed alternative is worth examining as it equals/outperforms the other alternatives, at least under certain circumstances. This is especially true in the presence of data scarcity and/or multiple contexts that make the training data incomplete hence inducing a high risk of bad generalization. 
\e Finally recall that in all circumstances, the proposed framework shows orders of magnitude shorter computation times and number of active coefficients and shows high potential for easier explainability and adoption in the industrial context.

\bibliographystyle{plain}  % Elsevier Harvard style for references
\bibliography{redundant_pwp}            % Path to your .bib file

\end{document}